\documentclass[conference]{IEEEtran}
\IEEEoverridecommandlockouts
\usepackage{cite}
\usepackage{amsmath,amssymb,amsfonts}
\usepackage{algorithmic}
\usepackage{graphicx}
\usepackage{textcomp}
\usepackage{xcolor}
\usepackage{stfloats}

\usepackage{cuted}
\usepackage{graphicx}
\usepackage{caption}
\usepackage{hyperref}
\hypersetup{hidelinks}
\usepackage{url}

\usepackage[font=footnotesize]{caption}
\usepackage{eso-pic}

\newcommand{\arxivnotice}{%
  \AddToShipoutPictureBG*{%
    \AtPageLowerLeft{%
      \put(\LenToUnit{0.5\paperwidth},\LenToUnit{0.35in}){%
        \makebox[0pt][c]{\fbox{\parbox{0.85\paperwidth}{\scriptsize\centering
          \copyright\,2026 IEEE. Personal use of this material is permitted. Permission from IEEE must be obtained for all other uses, in any current or future media, including reprinting/republishing this material for advertising or promotional purposes, creating new collective works, for resale or redistribution to servers or lists, or reuse of any copyrighted component of this work in other works.\\
          This is the accepted version of a paper to appear in the 2026 IEEE International Workshop on Metrology for Agriculture and Forestry (MetroAgriFor). The final published version will be available in IEEE Xplore.}}}}}}}

\def\BibTeX{{\rm B\kern-.05em{\sc i\kern-.025em b}\kern-.08em
    T\kern-.1667em\lower.7ex\hbox{E}\kern-.125emX}}

\title{ADA 2025: A Multimodal Dataset\\ for Water-Deficit Stress Detection in Sugar Beets}

\author{
  \IEEEauthorblockN{
    Piper Powell\IEEEauthorrefmark{1},
    Robert Franz\IEEEauthorrefmark{2},
    Amos Smith\IEEEauthorrefmark{1},
    Isaak Ihorst\IEEEauthorrefmark{1},
    Joachim Hertzberg\IEEEauthorrefmark{1}\IEEEauthorrefmark{2},
    Martin Atzmueller\IEEEauthorrefmark{1}\IEEEauthorrefmark{2}
  } 
  \IEEEauthorblockA{
    \IEEEauthorrefmark{1}\textit{Institute of Computer Science, Osnabrück University},
    Osnabrück, Germany \\
    \{piper.powell, amos.smith, isaak.ihorst, joachim.hertzberg, martin.atzmueller\}@uni-osnabrueck.de
  }
  \IEEEauthorblockA{
    \IEEEauthorrefmark{2}\textit{German Research Center for Artificial Intelligence (DFKI)},
    Osnabrück, Germany \\
    robert.franz@dfki.de
  }
}

\begin{document}
\bstctlcite{IEEEexample:BSTcontrol}
\arxivnotice

\maketitle

\begin{strip}
    \centering
    \includegraphics[width=0.71\textwidth]{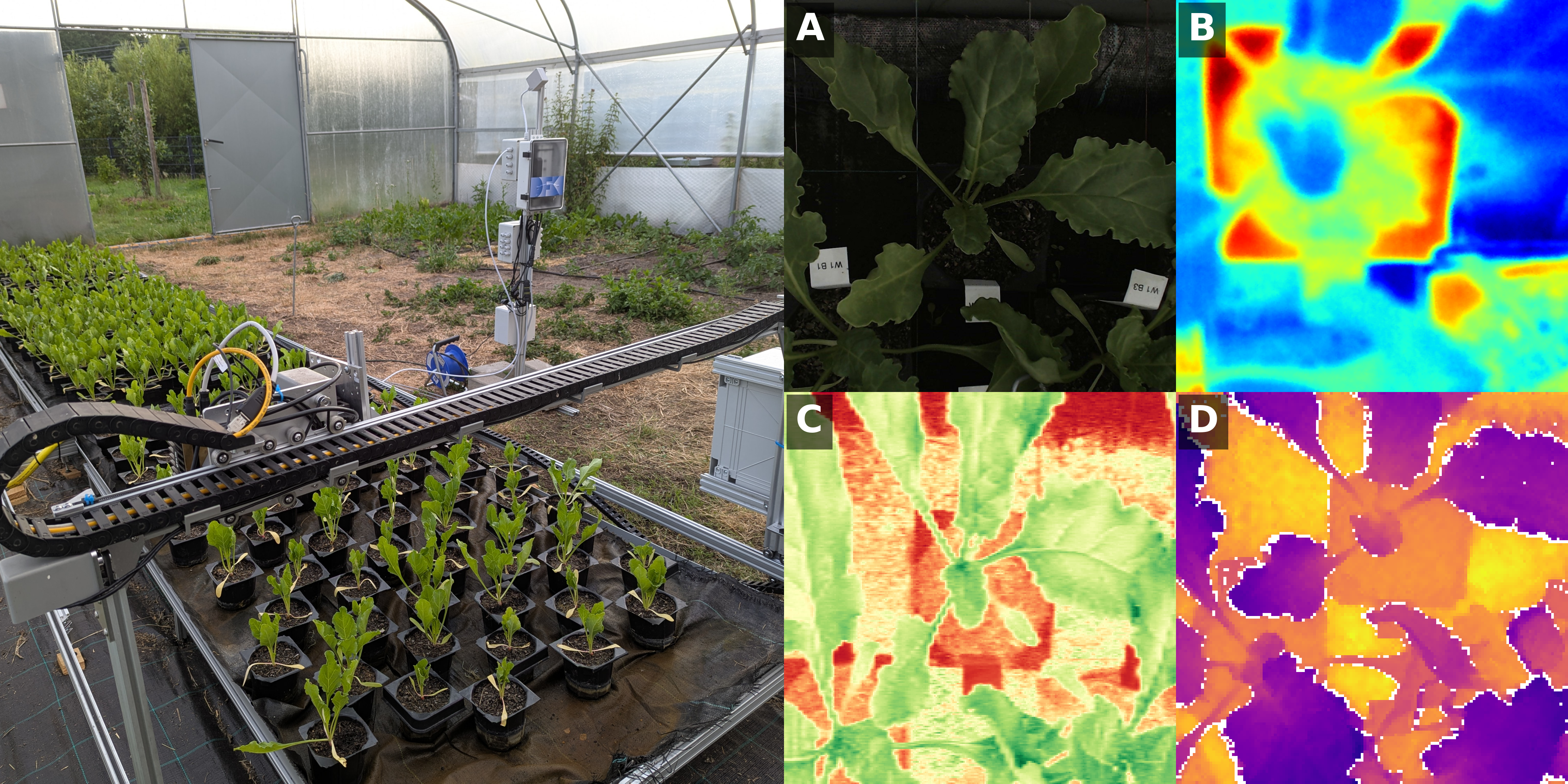}
    {%
    \captionsetup{width=0.71\textwidth}%
    \captionof{figure}{Overview image of the ADA robot and plant beds for data collection (left) and example images from the onboard RGB (A), thermal (B), hyperspectral (C), and depth (D) cameras.}%
    \label{fig:teaser}%
    }
\end{strip}

\begin{abstract}
Research and development into artificial intelligence (AI) and machine learning (ML) applications in crop management are hindered by a lack of suitable public datasets. Whereas applications in disease and pest detection benefit from a comparatively higher amount of available data, applications like water-deficit stress detection often have very few if any public datasets available. Moreover, most datasets that are available are single-modality and frequently RGB-only, leaving the early detection and enhanced performance offered by other methods, and particularly by multimodal approaches, out of reach. Datasets with accurate metadata, higher numbers of plants, and smaller intervals between data points are also scarce. To address this, we publicly release ADA 2025, a $>$500~GB comprehensive multimodal dataset of young sugar beet plants under healthy and water-deficit conditions combining thermal, hyperspectral, RGB, and depth imagery along with relevant environmental and metadata and both relative water content (RWC) and leaf thickness as ground truth sources. Exemplary results showcase the dataset's validity in terms of differences detectable between stressed and healthy plants.

\end{abstract}


\section{Introduction}

Growing pressures from climate change, limited resources, workforce depletion, and the loss of institutional knowledge resulting from a shortage of skilled workers have accelerated the adoption of technological solutions in crop management. In recent years, AI solutions for crop monitoring have gained increasing prominence in both practical applications and in research \cite{elbasi_artificial_2023,akhter_digital_2024,kisliuk2023ai}. With the impact of diseases and water scarcity expected to be critical challenges in the coming years, AI methods for the detection of plant diseases (e.g.,~\cite{khan_advancing_2024}) and water-deficit stress (e.g.,~\cite{kamarudin_deep_2021}) have seen particular growth. As with most of the broader AI sector, however, both applications face a substantial data scarcity problem, with water-deficit stress detection suffering from a particular lack of suitable datasets.

In contrast to general computer vision (CV) tasks, which are supported by widely adopted datasets such as MS-COCO \cite{lin2014microsoft} and ImageNet \cite{deng2009imagenet}, crop monitoring applications have access to comparatively few datasets. This scarcity stems largely from the difficulties of plant data collection, an inherently seasonal and fragile endeavor \cite{min_computer_2025}. Among the relatively few established crop monitoring datasets, the majority focus on disease or pest detection and rely exclusively on RGB imagery, including the widely used PlantVillage \cite{hughes_open_nodate} and the subsequent PlantDoc \cite{singh2020plantdoc} and FieldPlant \cite{moupojou_fieldplant_2023} datasets. Furthermore, datasets like PlantDoc incorporate internet-scraped images that frequently lack the metadata required to contextualize visible symptoms. These limitations are especially restrictive for water-deficit stress monitoring, for which RGB-only datasets are of limited use and contextual information is vital to differentiate ailments with visually similar symptoms. 

Water-deficit stress detection strongly favors technologies capable of detecting early symptoms and changes before visible damage (and resultant yield reduction) has occurred \cite{moustaka_early-stage_2023}, but data for many of these modalities is scarce. Hyperspectral images, for example, allow for the computation of various spectral indices related to water content, stress, and numerous other physiological variables \cite{zhang_estimation_2019} and have proven highly effective in stress detection (e.g.,~\cite{alordzinu_ground-based_2021}), but are greatly underrepresented in available public data. Similarly, thermal and depth data, which can provide early indications of stress based on leaf temperature changes \cite{risso_mems_2021} and leaf wilt \cite{su_evaluating_2019} respectively, are drastically underrepresented.

Although multimodal approaches consistently outperform single-modality methods in terms of predictive accuracy and robustness \cite{gao_fusion_2021}, datasets combining multiple sensing modalities remain exceedingly rare. To the best of the authors' knowledge, the only publicly available dataset suitable for context-aware, multimodal stress detection is the Eschikon Plant Stress Phenotyping Dataset developed by ETH Zurich \cite{khanna_spatio_2019}, which includes hyperspectral, depth, and RGB imagery. However, though a pioneering contribution, this dataset exhibits several important limitations. Each stress condition contains 18 plants across three replicate boxes, a reasonable but still modest sample size given the substantial inter-plant variability typically observed in phenotyping studies. Furthermore, data were collected only twice per week, preventing the capture of stress-induced changes in the natural diurnal and nocturnal physiological rhythms of the plants. 

To address the scarcity of datasets suitable for state-of-the-art multimodal water-deficit stress monitoring, we present and publicly release ADA 2025 \cite{ada2025_dataset}, a dataset comprising hyperspectral, RGB, depth, and thermal imagery together with environmental measurements, metadata, and ground-truth data. ADA 2025 is intended to support the development and evaluation of advanced AI-based water-deficit stress monitoring methods while also enabling broader investigations into plant responses to water-deficit stress and related secondary conditions in sugar beet crops.

\section{Method}

\subsection{Plants}
\begin{figure}[t]
  \centering
  \includegraphics[width=1.0\linewidth]{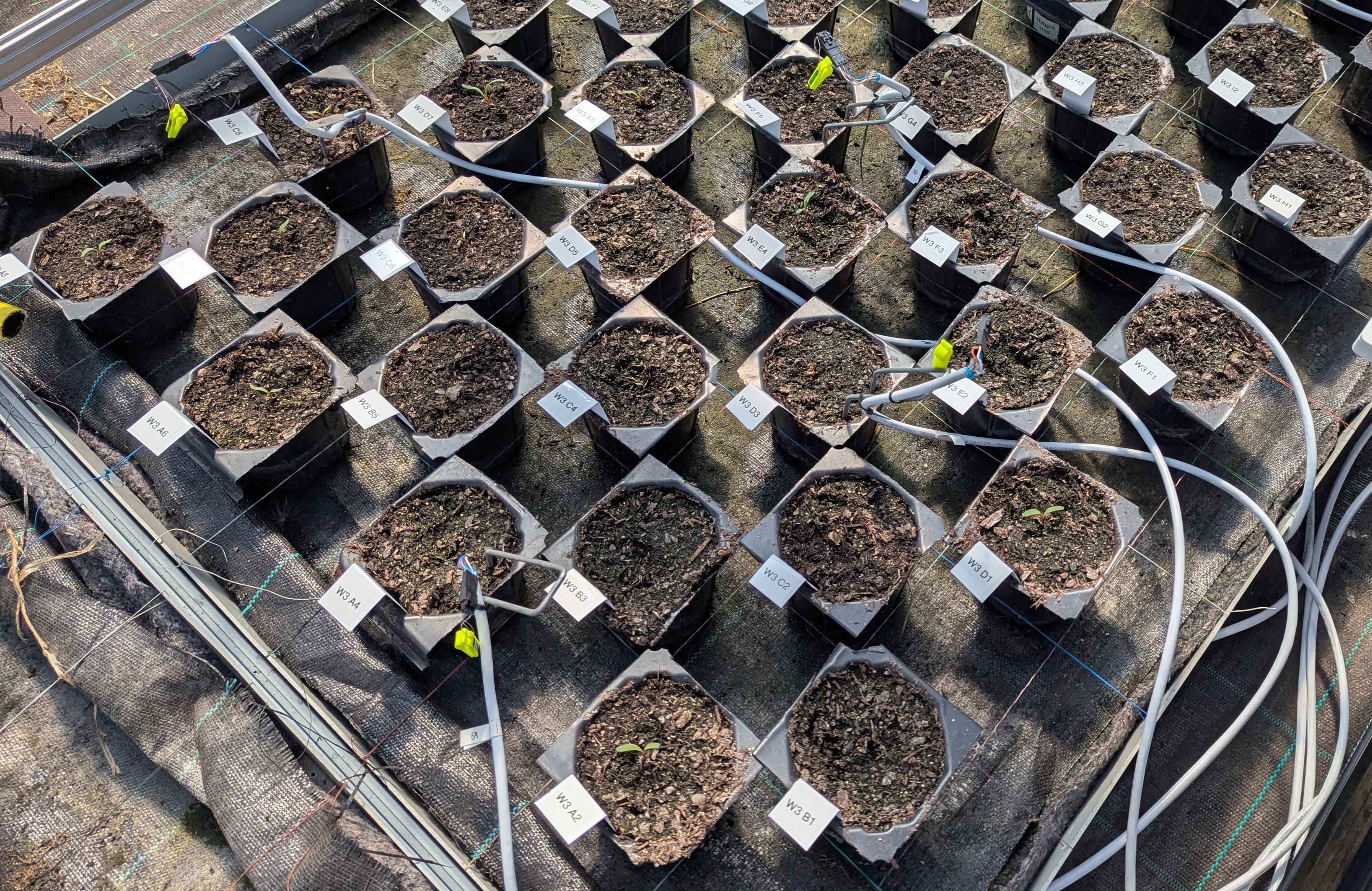}
  \caption{Overview of plant grid in one bed. Leaf sensors can be seen prepared at certain positions, awaiting attachment once the plants achieve the target growth stage.}
  \label{fig:plant_grid}
\end{figure}

A total of 216 sugar beet plants (\textit{Beta vulgaris ssp. vulgaris var. altissima}) of the variety \textit{CALLEDIA KWS Öko} were grown in individual pots in the greenhouse of the AI Market Garden of the German Research Center for Artificial Intelligence (DFKI) located in Bohmte, State of Niedersachsen during the 2025 growing season. All plants were seeded out on August~4 and then arranged in a checkerboard pattern (see Figure~\ref{fig:teaser} and Figure~\ref{fig:plant_grid}) in three plant beds of dimensions 2m x 1m, with an even division of plants between the beds, resulting in a total of 72 plants per bed. 

By September~3, the plants had reached a growth stage of approximately BBCH 14 (individual plants varied slightly) and irrigation to two of the beds was turned off, with the other bed maintained at 100\% watering. The plants in the former two beds then served as the water-deficit stress group and were allowed to dry out over the course of the collection period, while the plants in the remaining bed were maintained as healthy controls (beds referred to hereafter as the treatment and control beds respectively). The stress phase continued through September~15, by which time the plants in the treatment beds exhibited severe water-deficit stress symptoms. Water to the treatment beds was then turned back on to evaluate the final condition of the plants and to assess recovery potential. All but 3 plants were able to recover, albeit with substantial physical damage. Of the three non-surviving plants, two expired due to water-deficit stress. Collection concluded on September~22, at which time all plants were composted. 

\subsection{Platform}

The dataset was collected using the DFKI's ADA (Agricultural Data Acquisition) robot (see Figure~\ref{fig:teaser}). The base of the robot is the commercially available FarmBot Genesis system (v1.6), modified in-house to replace the standard Z-axis action arm with a camera platform and to extend its physical operating range over the full 6m x 1m experiment area of the current study. To handle the increased sensor load and to allow for more advanced control, the ADA robot was additionally upgraded with an Intel NUC small form factor PC and operated with ROS2.

During the collection period, the ADA robot performed flyovers of all three beds five times daily (6am, 9am, 12pm, 3pm, 6pm), capturing top-down views from all four cameras. This schedule was chosen to allow the capture of a representative sample of the full daylight period while balancing data capacity constraints. Data collection with the cameras began on August~12, 8 days post-seeding (emergence stage). The published dataset therefore spans the period from August~12 to September~22 (approx. 42 days). 

\subsection{Cameras}

\begin{figure}[t]
  \centering
  \includegraphics[width=1.0\linewidth]{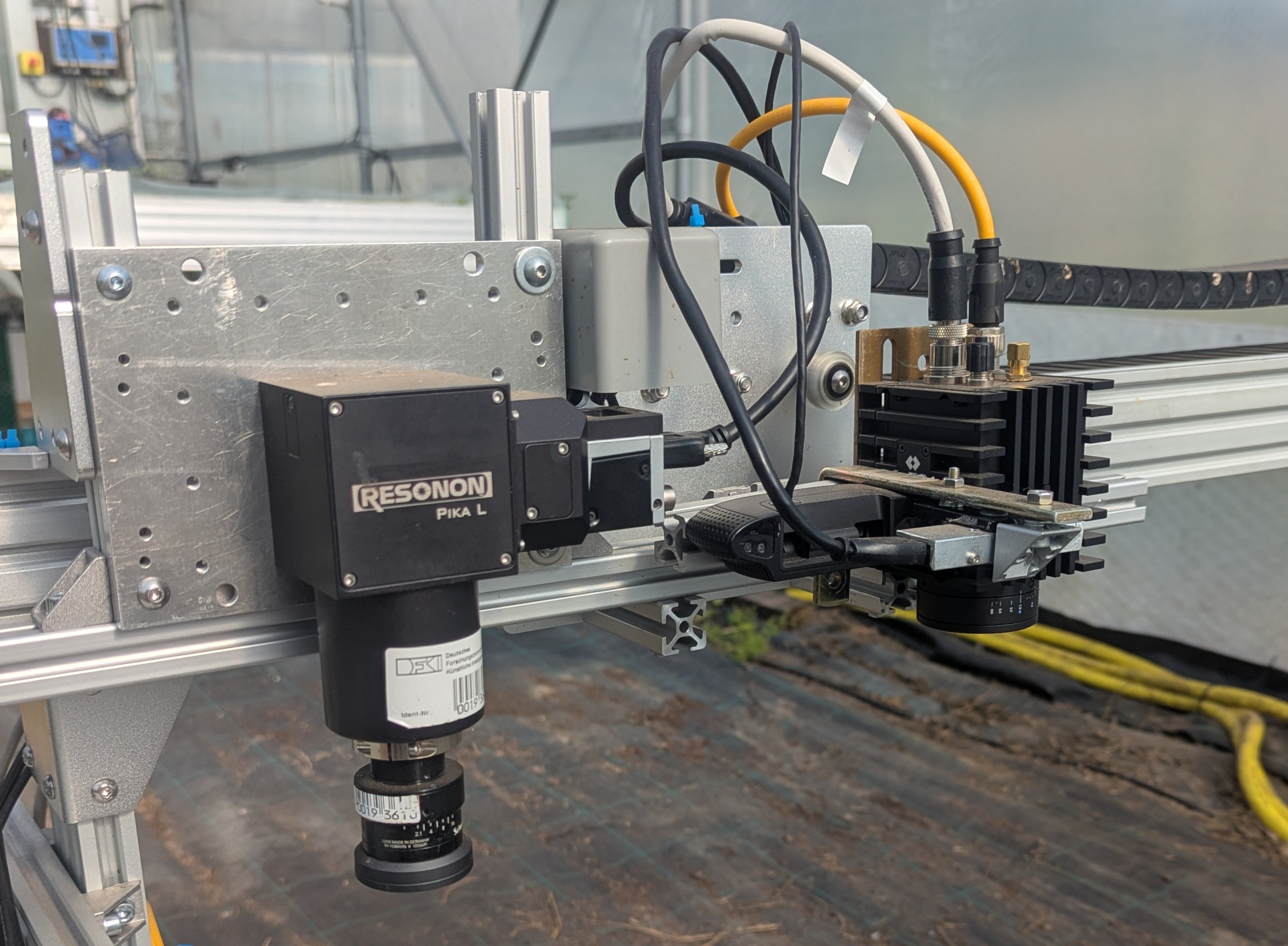}
  \caption{The camera package of the ADA robot, with the Pika L hyperspectral camera on the left, the A70 thermal camera on the right, and the Pico Flexx depth camera and Logitech webcam in the center, proximal and distal to the viewer respectively.}
  \label{fig:camera_pkg}
\end{figure}

Four cameras were selected for their established use in non-invasive stress detection (see Figure~\ref{fig:camera_pkg}): a Resonon Pika L hyperspectral camera (range of 400nm to 1000nm with channels in 2.7nm increments, resolution 900 x 1 pixels), a FLIR A70 thermal camera (accuracy ±2°C in the relevant~$-20$°C to 100°C range, resolution 640 x 480 pixels), a Pico Flexx depth camera (0.1--4 m range, depth resolution $\leq 1\%$ of distance (0.5--4 m @ 5fps) or $\leq 2\%$ of distance (0.1--1 m @ 45fps), resolution 224 x 171 pixels), and a standard C920 Logitech webcam. An overview of recording frequency and total image count for each camera can be found in Table \ref{tab:dataset}.

\subsection{Calibration}
\subsubsection{Geometric Calibration/Fusion}
As the cameras were mounted side by side, the raw images are not pixel-registered (cf.\ the example views in
Fig.~\ref{fig:teaser}). In lieu of markers or ground control points, joint calibration was achieved by using a chessboard to estimate the extrinsics of the RGB and thermal cameras relative to the depth camera by bundle adjustment (reprojection errors of $1.11$/$1.55$/$1.23$~px for RGB/thermal/depth). Undistortion was applied and the depth data were projected into the RGB and thermal frames to attach color and temperature values, accounting for parallax via the measured depth of each point. Because the hyperspectral camera is a push-broom sensor, its data could not be included in the fusion and were instead georeferenced using the robot's position log and stitched into per-row ortho-images. Calibration parameters and code are available in the code repository~\cite{ada2025_code}.

\subsubsection{Hyperspectral Calibration}
The hyperspectral images were flat-field-corrected using a dark frame $D$ and a 50\,\% gray reference
panel. The reference spectrum $W$ is the average response over the panel
region, captured once per plant row in each run and stored with every plant
image of that row. Each raw frame $R$
was corrected as $C = (R-D)/(W \cdot m - D)$ with $m=2.0$, so the gray panel
maps to $C\approx0.5$.

\subsection{Environmental Data}
We also include air temperature, humidity, pressure, and photosynthetically active radiation (PAR) measurements. Air readings were captured with two custom-built sensors using BME280 chips operated by an Arduino microcontroller. PAR data was collected using an industrial Driesen PAR sensor. An overview of recording frequency and total measurement count for each measure can be found in Table \ref{tab:dataset}.

\begin{table}[t]
\caption{ADA 2025 Dataset Contents by Modality}
\label{tab:dataset}
\centering
\setlength{\tabcolsep}{4pt}
\begin{tabular}{llrr}
\hline
\textbf{Modality} & \textbf{Measurement} & \textbf{Rate} & \textbf{Total} \\
\hline
\multicolumn{4}{l}{\textit{Camera Data}} \\
\hline
RGB          & Color         & 15 img./plant/run  & 761,400 \\
Depth        & Depth, Intensity           & 15 img./plant/run  & 694,440 \\
Thermal      & Infrared         &  5 img./plant/run  & 231,373 \\
Hyperspectral & Spectral        &  1 img./plant/run  &  50,544 \\
\hline
\multicolumn{4}{l}{\textit{Sensor Data}} \\
\hline
Air  & Temp., Humidity, Pressure  & Every 2 min.  &  59,648 \\
Leaf & Turgor Pressure            & Every 2 min.  & 294,247 \\
PAR  & Light Intensity            & Every 15 min. &   4,484 \\
\hline
\end{tabular}
\caption*{\footnotesize Note that approximately 6.4\% of the planned camera data could not be collected due to technical issues with the robot or cameras, resulting in the totals above; where a scheduled run failed, a re-run within one hour was used in its place where available. The differing images/plant/run figures between modalities are due to the differing frame rates of the respective cameras.}
\end{table}

\subsection{Ground Truth}

\subsubsection{Relative Water Content (RWC)}

Relative Water Content (RWC), calculated from the fresh, turgid, and dried weight of leaf tissue, is a widely used and effective indicator of plant water status \cite{barrs1968determination,mullan2012leaf}. To assess plant water status during the water-deficit period in our study, leaf samples were collected from the second water-deficit stress bed (bed 3, W3) at 1, 3, 5, 6, 7, 8, 11, and 12 days after irrigation was discontinued. On each sampling day, eight plants were selected, ensuring that each plant was sampled only once over the course of the experiment, and the youngest, fully developed leaf was cut from each plant. Fresh weights were recorded immediately following collection. Leaves were then rehydrated at room temperature for 3 hours to obtain turgid weight and subsequently dried in a $70^\circ\mathrm{C}$ oven for 3 hours to obtain dry weight. Additional image scans of the fresh leaf samples are included in the dataset, enabling other metrics related to plant water content, such as the spectral-correlated Equivalent Water Thickness (EWT) \cite{MIAO2024111497}.

\subsubsection{Turgor Sensors (Leaf Clips)}

In all, 21 AgriHouse Leaf Sensor Rev4 clips were used across the three beds to capture changes in leaf thickness (a measure of turgor pressure), providing a high frequency and rapidly responding baseline for real time leaf water content  \cite{seelig_irrigation_2012}. Nine sensors each were deployed in a Z-shaped pattern in the treatment beds to provide a representative sample, with the remaining three sensors attached to plants down the center axis of the control bed to provide a clean healthy baseline. 

\subsubsection{Plant Observations}

In addition to the technical measurements of plant status, we also include a log of manual BBCH estimations and plant status and symptom observations from the experiment team for the period beginning with the inducement of water-deficit stress and continuing through to the end of collection, including relevant symptom images and a final set of images of all plants individually as they appeared on the final day of collection before composting. This additional metadata also includes notes on the pests and diseases observed in the experiment plants, namely the black bean aphid (\textit{Aphis fabae}) as well as powdery (\textit{Erysiphaceae}) and downy mildews (\textit{Peronosporaceae}), expected ailments given the stress induced in the plants and the time of year in which data collection occurred.

\section{Exemplary Applications and Results}
\begin{figure}[hbt]
    \centering
    \includegraphics[width=1.0\linewidth,trim={0 9 0 0},clip]{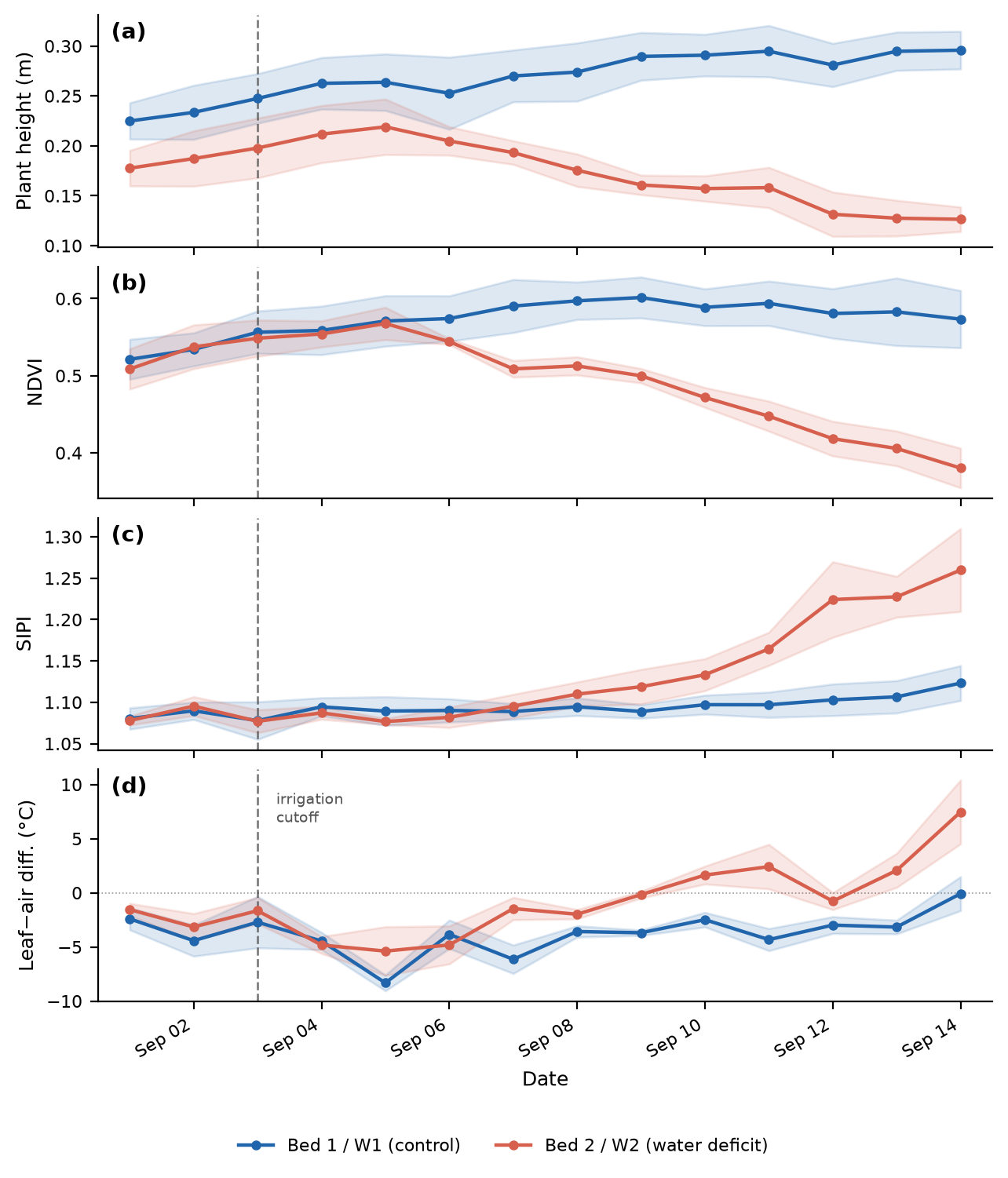}
    \caption{
    Trajectories of plant height, NDVI, SIPI, and leaf-air temperature over time in the control and treatment bed. The irrigation cutoff on September~3 is marked for reference. Shaded bands indicate ±1 standard deviation. Note that the 3pm thermal capture of the treatment bed failed on September~11; the reading shown for that day comes from a re-run at 4pm.
    }
    \label{fig:results}
\end{figure}

The code for the exemplary analyses below is available on GitHub \cite{ada2025_code}. All models and correlations use data from September 3 (irrigation cutoff) through September 14 (one day prior to irrigation resumption). Figure~\ref{fig:results} additionally shows September 1--2 to illustrate similar behavior in both beds prior to the cutoff. Data for the 3pm runs (approx. 1.5 hours after local solar noon) were used for these analyses to capture the behavior of the plants at the hottest part of the day at a consistent daylight hour well clear of both the morning and evening irrigation rounds (in the control bed). Five plants were chosen per bed (four corners and center) to cover both edge and central positions and yield a representative sample of the plants in those conditions. Analyses were performed on data from the control bed (bed 1, W1) and the first water-deficit stress bed (bed 2, W2), referred to as ``the treatment bed'' in this section. Data for the second water-deficit stress bed (bed 3, W3) was not included in the exemplary analyses, but is available in the full dataset.

\subsection{Plant Height Estimation from RGB-D Imagery}
RGB-D point clouds were generated from the synchronized RGB and depth images, which were undistorted using OpenCV's pinhole camera distortion model and then reconstructed using the calibrated camera parameters provided in the code repository \cite{ada2025_code}. All points belonging to plants were segmented from the background using the excess green index (ExG) \cite{woebbecke1995color} computed on the 8-bit RGB channels, with a threshold value of 5 chosen based on qualitative assessment of segmentation performance. Plant height was estimated as the 75th percentile of segmented vegetation points, a robust canopy height metric commonly used in remote sensing and phenotyping applications to reduce sensitivity to outliers and reconstruction noise \cite{hutt2023uav}.

Plant height was modeled using a linear mixed-effects model with day, treatment, and their day-by-treatment interaction as fixed effects, and a random intercept per plant to account for repeated measurements. The control plants exhibited a significant increase in estimated height over the study period ($\hat{\beta}_{\mathrm{day}} = 0.00430$~m/day, $p = 4.61 \times 10^{-14}$), while the cessation of irrigation significantly altered the temporal trajectory of height ($\hat{\beta}_{\mathrm{day:treatment}} = -0.01308$~m/day, $p = 3.45 \times 10^{-59}$). Estimated height in the treatment bed ceased increasing several days after irrigation was discontinued and subsequently declined, consistent with drought-induced wilting, cf. Panel (a) in Figure~\ref{fig:results} for a visual overview.

\subsection{Hyperspectral Indices and Leaf Temperature Analysis}
As the hyperspectral and thermal values were aggregated per bed, ordinary least-squares models were used, as each bed-date observation is independent with no location-based random effects. In both cases, the data were modeled with treatment, days since intervention, and their interaction as predictors. 

Three hyperspectral indices were calculated for this analysis: NDVI $= (R_{\mathrm{NIR}} - R_{\mathrm{Red}})/(R_{\mathrm{NIR}} + R_{\mathrm{Red}})$ with $R_{\mathrm{NIR}} \approx 800$nm, $R_{\mathrm{Red}} \approx 670$nm \cite{rouse1974monitoring}; CI Red Edge $= (R_{\mathrm{NIR}}/R_{\mathrm{RedEdge}}) - 1$ with $R_{\mathrm{NIR}} \approx 790$nm, $R_{\mathrm{RedEdge}} \approx 720$nm \cite{gitelson2003relationships}; and SIPI $= (R_{800} - R_{445})/(R_{800} - R_{680})$ \cite{penuelas1995semi}. These were chosen from the full set of 21 indices computed on the hyperspectral data as they represent standard, well-established indices that combined show both directions of index response to water-deficit stress (some indices rise as stress increases; others drop). For the hyperspectral indices, plant pixels were separated from the background using a segmentation model trained on 234 manually annotated masks created from the plants used in the analysis. Spectra were smoothed with a Savitzky-Golay filter (window length 31 bands, polynomial order 3), and each band value entering an index is the mean over a five-band window around the stated center wavelength. Index values were first averaged over the plant pixels of each image and then aggregated to daily per-bed means.

 In the treatment bed (W2), the treatment $\times$ time interaction was significant for all three indices: NDVI ($\hat{\beta}_{\mathrm{W2:day}} = -0.0189$/day, $p = 1.39 \times 10^{-9}$), CI Red Edge ($\hat{\beta}_{\mathrm{W2:day}} = -0.0098$/day, $p = 6.97 \times 10^{-6}$), and SIPI ($\hat{\beta}_{\mathrm{W2:day}} = +0.0143$/day, $p = 1.83 \times 10^{-6}$). NDVI and CI Red Edge declined in the treatment bed while remaining stable or slightly increasing in the control bed, consistent with progressive chlorophyll loss and reduced photosynthetic activity under water-deficit stress. SIPI showed the expected inverse response, increasing in the treatment bed while remaining stable in the control bed. See Panels (b) and (c) in Figure~\ref{fig:results} for a visual overview (note that because CI Red Edge behaves similarly to NDVI, only NDVI is shown in the figure).

The difference between leaf temperature and surrounding air temperature was determined by segmenting out the coolest 50\% of the pixels in the thermal images for each plant, calculating the median temperature of the leaf pixels, and comparing this to the median recorded air temperature value for the relevant period. Per-plant leaf-air temperature differences were aggregated to daily per-bed medians before modeling. The treatment $\times$ time interaction term indicated a significantly faster rate of leaf warming in the treatment bed relative to the control ($\hat{\beta}_{\mathrm{W2:day}} = +0.596$°C/day, $p = 0.030$). This pattern is consistent with the loss of evaporative cooling ability in the treatment plants, which ultimately achieved leaf temperatures above the ambient air temperature while the control plants remained at or below the ambient temperature level. See Panel (d) in Figure~\ref{fig:results} for a visual overview.

\subsection{Correlations with Turgor Ground Truth}
In the treatment bed, Spearman correlations were calculated between the turgor sensor readings and each of the four stress indicators (the three hyperspectral indices and the leaf-air temperature difference) over time. Turgor correlated significantly with all four: NDVI ($\rho = -0.930$, $p = 1.2 \times 10^{-5}$), CI Red Edge ($\rho = -0.860$, $p = 3.3 \times 10^{-4}$), SIPI ($\rho = 0.944$, $p = 3.9 \times 10^{-6}$), and leaf-air temperature difference ($\rho = 0.853$, $p = 4.2 \times 10^{-4}$). Since the turgor sensors output \textit{higher} readings as turgor pressure decreases (indicating increasing water-deficit stress), these correlation signs match expectations.

\section{Limitations}
As with any dataset, care must be taken with respect to a few key points when utilizing the ADA 2025 data. When segmenting the plants from the background longitudinally across the full time span of the dataset, attention must be paid to the shifting contents of the background as the plants grow and overlap both the pots and each other. While this is normal and reflective of real-world conditions any resulting models would have to face in live application, it does require the selection of a segmentation method not reliant on the precise contents of the background. The methods used in the exemplary analyses target the plant pixels via either a segmentation model trained on manually annotated masks (hyperspectral) or the plants' spectral and thermal properties (RGB-D and thermal). Notably, as low-water conditions also predictably influence the spectral, thermal, and physical responses of the surrounding background material (e.g., material heating and drying, soil shrinking), background pixels in this application represent additional supporting data useful for accurate state diagnosis rather than noise. A sensitivity check in the repository probes each modality's control--treatment difference over a range of segmentation thresholds, with the direction of the difference holding at every threshold in all four modalities. Further information and code for evaluating the influence of the background, and for effective segmentation, is available in the associated code repository~\cite{ada2025_code}.

As is the case with any large-scale dataset, there are also time points in the dataset where a sensor was down for a technical problem, where the capture conditions were not ideal, or where other transient issues affected the data. For example, images from the early and late runs can be dark in the RGB and hyperspectral data, and as the hyperspectral exposure is set automatically once per run when the sensor starts, changes in lighting during a run can lead to overexposed hyperspectral images. The cases where this occurred are documented in the dataset's associated code repository~\cite{ada2025_code}.

Notably, we do not position this dataset as a benchmark in the sense of a full ML pipeline. The statistical analyses show that the data provides a detectable stress signal in various modalities, but the full implementation of a model or pipeline for leveraging these signals in live detection is treated as out of scope for this introductory paper. Future work leveraging, e.g., regression and multimodal sensor fusion can build on the foundation provided by the dataset in the development of deployable systems for monitoring plant stress.

\section{Conclusion}
The ADA 2025 dataset fills a gap in existing data supply for cutting-edge crop management, providing a large multimodal dataset of high density and quality, suited to developing and evaluating a variety of AI- and ML-based plant monitoring methods. Although it is not guaranteed that all conclusions or models based on this data will apply to other sugar beet varieties or to crops grown outside of a greenhouse environment, the dataset nonetheless provides a foundation for continued development of general multimodal methods and models for water-deficit stress detection (as well as general plant phenotyping), supporting the ultimate creation of next generation systems able to efficiently monitor and manage crops.

Immediate avenues for future work include collecting data in the same paradigm for other plant species and extending the collection schedule to include multiple runs over the full growing season. The inclusion of soil moisture data, a key variable in the full water-deficit cycle, would also represent a substantive improvement (soil sensors were deployed in this study, but the sensors used ultimately proved unsuitable for the deployment conditions). Follow-up studies extending the growth period out to harvest would also allow for the collection of yield data, a factor of vital financial importance to farmers and growers. In addition, long-term extensions leveraging the dataset to train the necessary detection components of next-generation irrigation management systems would further aid the crop management sector in adapting to the challenges it faces in coming years.

\section*{Acknowledgment}
This work was supported by the Lower Saxony Ministry of Science and Culture (MWK), funded through the zukunft.niedersachsen program of the Volkswagen Foundation, as part of two projects (ZN4219 -- HybrInt and ZN4661 -- RLA). The DFKI Niedersachsen is supported by the Ministry of Science and Culture of Lower Saxony from funds of the program zukunft.niedersachsen of the Volkswagen Foundation. We also thank the agrifood-TEF-DE project for additional hardware support. AI tools (Claude, GitHub Copilot) were used to assist with analysis code and reporting, figure and table generation, and minor revisions to the text. All code and content was reviewed and checked by the authors.

\bibliographystyle{IEEEtran}
\bibliography{references}

\end{document}